# Graph Convolutional Long Short-Term Memory Attention Network for Post-Stroke Compensatory Movement Detection Based on Skeleton Data

Jiaxing Fan
School of Control Science and Engineering
Shandong University
Jinan, China
202334915@mail.sdu.edu.cn

Jiaojiao Liu
Rehabilitation and Physical Therapy Department
Shandong University of Traditional Chinese Medicine Affiliated Hospital
Jinan, China
1142365430@qq.com

Wenkong Wang
School of Control Science and Engineering
Shandong University
Jinan, China
202320760@mail.sdu.edu.cn

Yang Zhang*
Rehabilitation and Physical Therapy Department
Shandong University of Traditional Chinese Medicine Affiliated Hospital
Jinan, China
zhangyang982003@163.com

Xin Ma*
School of Control Science and Engineering
Shandong University
Jinan, China
maxin@sdu.edu.cn

Jichen Zhang*
Shandong Inspur Science Research Institute Co., Ltd.
Jinan, China
zhangjichen@inspur.com

*Abstract*—Most stroke patients experience upper limb motor dysfunction. Compensatory movements are prevalent during rehabilitation training, which is detrimental to patients' long-term recovery. Therefore, detecting compensatory movements is of great significance. In this study, a Graph Convolutional Long Short-Term Memory Attention Network (GCN-LSTM-ATT) based on skeleton data is proposed for the detection of compensatory movements after stroke. Sixteen stroke patients were selected in the research. The skeleton data of the patients performing specific rehabilitation movements were collected using the Kinect depth camera. After data processing, detection models were constructed respectively using the GCN-LSTM-ATT model, the Support Vector Machine(SVM), the K-Nearest Neighbor algorithm(KNN), and the Random Forest(RF). The results show that the detection accuracy of the GCN-LSTM-ATT model reaches 0.8580, which is significantly higher than that of traditional machine learning algorithms. Ablation experiments indicate that each component of the model contributes significantly to the performance improvement. These findings provide a more precise and powerful tool for the detection of compensatory movements after stroke, and are expected to facilitate the optimization of rehabilitation training strategies for stroke patients.

## I. Introduction

Stroke is one of the diseases with the highest mortality rate. Up to 80% of stroke patients suffer from upper limb motor function disorders of varying degrees, which seriously affect their daily activities[1]. Rehabilitation training after stroke is crucial for patients to regain their upper limb motor function. During the rehabilitation process, in order to achieve the intended actions, patients often change their movement patterns and use other joints or body parts to compensate for the loss of limb motor function, which is an important cause of compensatory movements[2][3]. The movement of the upper limbs is extremely complex. Compensatory movements mainly include trunk lean-forward(TLF), trunk rotation(TR), and shoulder elevation (SE)[4]. As the rehabilitation progresses, these compensatory movements are not conducive to the correct and effective rehabilitation of the affected side. In daily life, compensatory movements can help improve the short-term outcomes of patients. However, long-term compensatory movements will hinder the affected side from returning to the pre-stroke movement state and may even lead to the occurrence of other diseases[5]. Therefore, it is of great significance to detect these compensatory movements during the rehabilitation process of patients and provide guidance and reminders to reduce such compensatory movements.

Currently, clinical methods for reducing compensatory movements largely rely on the supervision, guidance, and correction of professional rehabilitation doctors, which is both time-consuming and labor-intensive[6]. From the perspective of doctors, they are burdened with heavy workloads and have limited time and energy. There is an urgent need for an automatic detection of compensatory movements during the rehabilitation process, eliminating the necessity of one-on-one manual assistance from doctors. From the perspective of patients, there is also an urgent need for automatic detection of compensatory movements during the rehabilitation process, especially for those who are undergoing home-based rehabilitation treatment. Such information can assist doctors and patients in formulating more strategic rehabilitation strategies.

In the current research, the utilization of advanced devices and algorithms has become a crucial direction in the field of upper limb compensatory movement detection. At present, the mainstream detection methods in this field rely on devices such as Inertial Measurement Units(IMUs)[5][7], pressure pad sensors[8], surface electromyography(sEMG) sensors[9][10], depth cameras[11], and the Vicon optical tracking system[12]. Among them, pressure-based detection methods are vulnerable to external environmental interference, resulting in unstable pressure collection[8]. The detection solutions based on wearable devices suffer from issues such as cumbersome installation and poor stability[4]. Moreover, the detection methods based on the Vicon optical

Jiaxing Fan, Jiaojiao Liu—Authors with equal contributions.
National Key Research and Development Program Project under Grant 2023YFB4706104. Key R&D Program of Shandong Province (Major Science and Technology Innovation Project) under Grant 2024CXGC010603. Fundamental Research Funds for the Central Universities under Grant 2022JC011.

tracking system face challenges including high costs, the need for marking workpieces, and complex operation procedures[12].

In contrast, the Kinect depth camera sensor stands out due to its unique advantages. As a non-invasive, non-wearable, and more cost-effective platform, the Kinect sensor has received significant attention in the field of human motion detection[13]. It is capable of achieving real-time 3D human skeleton tracking, effectively reducing the impact of differences in human appearance and lighting changes on the detection results. At the same time, deep learning algorithms can accurately analyze the skeleton data collected by the Kinect and precisely identify the corresponding human movements, providing an efficient and reliable technical approach for the detection of upper limb compensatory movements[14]. Therefore, the method of detecting compensatory movements based on skeleton data has emerged as a viable solution. The main contributions of this paper are as follows:

1、A method for detecting compensatory movements based on contactless skeleton data, namely the Graph Convolutional Long Short-Term Memory Attention Network, is proposed.

2、Through comparative experiments, it is verified that in the task of detecting compensatory movements, the deep learning algorithm proposed in this paper achieves a higher accuracy rate compared with the existing three commonly used machine learning methods, effectively demonstrating the effectiveness of the proposed method.

## II. METHODS

In this study, a method for detecting compensatory movements based on skeleton data is proposed, and the specific process structure of this method is shown in Figure 1. Firstly, the subjects are required to complete rehabilitation exercises under the monitoring of the Kinect depth camera, so as to collect the original movement data of the subjects. Subsequently, the collected data undergoes operations such as preprocessing, skeletal point extraction, and feature extraction in sequence. Finally, the deep learning algorithm proposed in this study and three existing machine learning algorithms are respectively employed to construct a compensatory movement detection model, achieving accurate identification and analysis of compensatory movements. In Figure 1, NC=Noncompensation, TLF=Trunk lean-forward, TR=Trunk rotation, SE=Shoulder elevation.

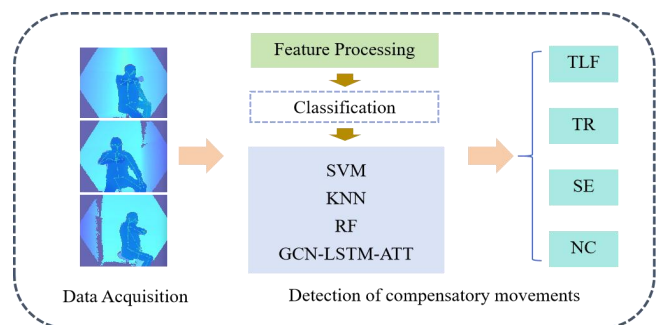

Figure 1. Structure of the Workflow

### A. participants

The subjects included 15 stroke patients recruited from the Affiliated Hospital of Shandong University of Traditional Chinese Medicine. The subjects provided written informed consent before the experiment, and the relevant procedures and experimental protocols were approved by the Institutional Ethics Committee of the Affiliated Hospital of Shandong University of Traditional Chinese Medicine.

The inclusion criteria for stroke patients are as follows: 1) Meeting the diagnostic criteria for stroke and confirmed by CT or MRI; 2) Aged between 20 and 80 years old; 3) Patients who have experienced their first stroke more than 15 days ago and within 6 months; 4) At Brunnstrom stage Ⅱ or above; 5) Having a clear understanding of the research protocol and being able to comply with it (MMSE>24 points); 6) Volunteering to participate and signing the informed consent form.

The exclusion criteria are as follows: 1) Severe cognitive impairment that hinders the completion of the Fugl-Meyer Assessment (FMA) scale test; 2) Other physical injuries, such as fractures, severe arthritis, etc.; 3) Severe insufficiency of heart, liver, and kidney functions; 4) Other situations that the researchers consider unsuitable for participating in this experiment.

### B. Experimental System and Process

In the dataset collection system, three Kinect cameras placed at different positions are used to capture the 3D skeletal point data of the participants. The Kinect records depth videos with a resolution of 320 × 288 pixels and a working frequency of 30Hz. The operating range of the Kinect cameras is 0.5-5.46m. The 3D skeleton data is transmitted from the depth videos by the corresponding Software Development Kit (SDK). An overhead view of the placement of the data collection equipment is shown in Figure 2.

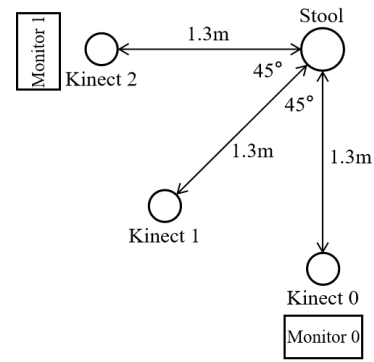

Figure 2. Overhead View of the Placement of Data Collection Equipment

The adoption of a multi-view data collection strategy is more in line with the practical needs of compensatory movement detection. This deployment method can comprehensively capture the movement information of the subjects, effectively avoiding the potential problem of information loss that may occur in a single view. In addition, the experiment requires each stroke subject to reach their maximum possible range of motion when performing rehabilitation exercises, so as to fully expose potential compensatory movement behaviors. Regarding whether the subjects exhibit compensatory movements, this study employs a dual-expert independent evaluation mechanism. Two experts with rich clinical experience conduct evaluations according to professional standards. The final judgment is reached through cross-checking and discussion, thus ensuring the accuracy and reliability of the evaluation results.

(1)Experimental Task Design: Based on the suggestions of clinical rehabilitation doctors and relevant research works, this study has carefully selected three standard rehabilitation movements that are likely to trigger compensatory movements during the daily rehabilitation training of stroke patients. Meanwhile, three common compensatory movements corresponding to them have been determined. The specific details and illustrations of these movements are presented in Table 1 and Figure 3 respectively, so as to clearly show the content of the experimental design. All the movements are derived from the routine rehabilitation training programs formulated by clinical rehabilitation doctors for stroke patients. The reason for selecting these movements is that they have a relatively high incidence of compensatory movements in clinical practice.

Table 1. Standard Rehabilitation Movements and Corresponding Compensatory Movements

|   | Rehabilitation actions | Corresponding compensatory movements |
|---|---|---|
| 1 | Touch one's mouth | Trunk lean-forward |
| 2 | Extend backward | Trunk rotation |
| 3 | Abduction of the arm | Shoulder elevation |

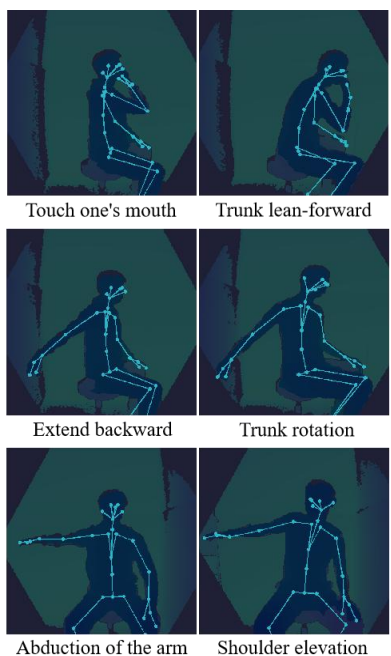

Figure 3. Standard Rehabilitation Movements and Corresponding Compensatory Movements

(2) Experimental Setup: In this experiment, the subjects are required to sit on a fixed round stool, and the motion data is synchronously collected from different perspectives by three Kinect depth cameras. The subjects need to complete three rehabilitation movements, namely touching the mouth, abducting the upper limb by 90 degrees, and extending it backward by 45 degrees, in sequence, at a comfortable movement speed and within the maximum range of motion they can achieve. Before the experiment starts, the subjects will be fully acquainted with the experimental procedure and the movement specifications under the guidance of professional doctors.

During the experiment, each subject is required to repeat each rehabilitation movement six times. There is an interval of 8 seconds between each single movement, and a rest period of 1 minute is set between adjacent movements to avoid the interference of fatigue factors on the authenticity of the experimental data. Throughout the experiment, two experienced clinical rehabilitation doctors closely observe the subjects and record in real time the occurrence of three common compensatory movements: trunk forward lean, trunk rotation, and scapula elevation.

After the experiment, two senior doctors independently assess the upper limb motor function of each subject using the upper limb part of the Fugl-Meyer Assessment Scale (FMA-UE). Finally, the average of the scores given by the two doctors is taken as the quantitative score of the subject's upper limb motor function.

C. *Skeleton Data and Preprocessing*

(1)Skeleton Data: In this study, the Azure Kinect Body Tracking SDK is used to extract the three-dimensional motion data (X, Y, Z) of 32 skeletal joint points from the depth video. In the analysis of upper limb movements based on the skeleton, many methods only use the coordinates of the upper limb joints as the input. This strategy is implemented to prevent the introduction of redundant information from additional nodes[15][16]. In this work, the skeleton data is composed of the coordinates of 20 joints of the upper limb. At the spatial dimension level, the joints within each frame are orderly connected according to the human anatomical structure, thus forming a complete topological structure of the upper limb skeleton. The specific connection method is shown in Figure 4.

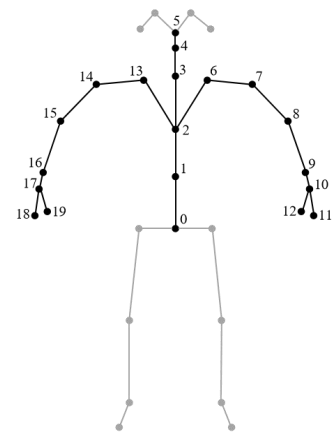

Figure 4. Spatial Skeleton Diagram

(2)Data Preprocessing: Firstly, the key frames of the original action sequence are intercepted. The Euclidean distance of the position change of the skeletal points between adjacent frames is calculated, and the frames whose distance exceeds the set threshold are retained. Subsequently, the sliding window algorithm is used for further processing. The window size and step length are set, and the similar and redundant frames within the window are removed, so as to retain the complete and effective action segments. To address the issue of differences in the duration of action sequences, this study proposes a time-axis normalization strategy based on the longest sequence, and combines it with the cubic spline interpolation technique to achieve accurate temporal alignment of different samples and ensure the consistency of all data in the time dimension. At the same time, the data reliability is further enhanced by performing Z-score normalization processing on the three-dimensional coordinates.

*D. Graph Convolutional Long Short-Term Memory Attention Network (GCN-LSTM-ATT)*

The Graph Convolutional Long Short-Term Memory Attention Network method proposed in this study is a deep learning model that organically integrates the advantageous characteristics of the Graph Convolutional Network (GCN), the Long Short-Term Memory network (LSTM), and the Attention mechanism (ATT). An overview of this model is shown in Figure 5.

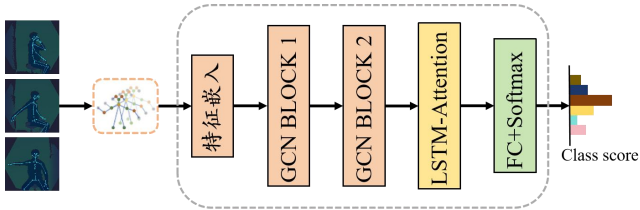

Figure 5. Graph Convolutional Long Short-Term Memory Attention Network

In terms of structural principles, this model regards the human body skeleton as a unique graph structure and makes full use of the powerful graph data processing ability of the Graph Convolutional Network (GCN). In the modeling of human skeletal joint points, GCN can effectively capture and analyze the complex spatial dependency relationships among various joint points. Through operations on the nodes and edges in the graph structure, GCN can extract the spatial correlation features among the joint points, thus accurately describing the characteristic distribution of human actions in the spatial dimension[17].

The Long Short-Term Memory network (LSTM) plays an important role in processing time series data within this model. In the scenario of human action analysis, action sequences exhibit obvious characteristics of temporal dynamic changes. Through its unique gating mechanism, LSTM can effectively learn and remember the dependencies of action sequences at different time steps[18]. Whether it is the starting, ongoing, or ending stage of an action, LSTM can accurately capture the key information in the time series, avoiding the common problems of gradient vanishing and gradient explosion in traditional Recurrent Neural Networks (RNNs), thus enabling stable modeling of the temporal dynamics of actions.

The introduction of the Attention mechanism (ATT) further enhances the performance of the model. When processing action sequences, the information at each time step does not have the same level of importance. ATT can adaptively adjust the degree of attention paid to different time steps according to the characteristics of the input data. Specifically, it can intelligently identify the key frames in the action sequence and assign higher weights to these key frames, thereby highlighting the significance of the key information. At the same time, for the time steps that may contain noise or relatively unimportant information, ATT will suppress their weights to reduce the impact of noise interference. In this way, the Attention mechanism significantly improves the discriminative ability of the model, enabling it to classify and identify actions more accurately.

In conclusion, by integrating GCN, LSTM, and ATT, the Graph Convolutional Long Short-Term Memory Attention Network (GCN-LSTM-ATT) can not only accurately model the spatial dependency relationships among human skeletal joint points, but also effectively capture the temporal dynamic changes of action sequences. Additionally, it can adaptively increase the weights of key frames and suppress noise interference. It has demonstrated powerful performance and potential in the field of human action analysis.

*E. Machine learning algorithms*

In conclusion, by integrating GCN, LSTM, and ATT, the Graph Convolutional Long Short-Term Memory Attention Network (GCN-LSTM-ATT) can not only accurately model the spatial dependency relationships among human skeletal joint points, but also effectively capture the temporal dynamic changes of action sequences. Additionally, it can adaptively increase the weights of key frames and suppress noise interference. It has demonstrated powerful performance and potential in the field of human action analysis.

1）SVM

The Support Vector Machine (SVM) is a classic supervised machine learning algorithm. Its core lies in finding the optimal hyperplane that can maximize the class interval for data classification[19]. When the data is linearly separable, the hyperplane can be directly solved. When the data is linearly inseparable, the kernel function is used to map the data into a high-dimensional space for processing. Its advantages include strong learning ability with small samples, good adaptability to high-dimensional data, and excellent generalization performance. However, its limitations are as follows: it has high computational complexity when dealing with large-scale data; it is originally designed for binary classification, and extending it to multi-class classification is rather cumbersome; and it is sensitive to the distribution of data.

2）KNN

The K-Nearest Neighbors algorithm (KNN) is a simple and intuitive supervised machine learning method. Its core logic is as follows: by calculating the distances (commonly using metrics such as Euclidean distance) between the data point to be classified and each point in the training set, the K nearest neighbors are found, and the category with the highest frequency among these K neighbors is taken as the category of the data point to be classified. The advantages of KNN are that it does not require complex training and can adapt to various data distributions. However, its disadvantages are that it needs to traverse the training set during calculation, resulting in low efficiency in scenarios with large amounts of data. Moreover, it is extremely dependent on the selection of the K value. An inappropriate K value will significantly affect the classification accuracy[20].

3）RF

Random Forest (RF) is an ensemble learning algorithm built based on decision trees. It generates sub-datasets through bootstrap sampling from the training set, constructs a decision tree for each sub-dataset, and randomly selects a part of the features to determine the splitting points during the construction process. In the case of classification, the result is determined by voting, while in regression, the mean value is taken. Its advantages include high accuracy, strong robustness, the ability to handle high-dimensional data, and a certain degree of interpretability. However, its disadvantages are high computational complexity and difficulty in comprehension[21].

## III. EXPERIMENTAL RESULTS

In this section, based on the data collected from real subjects, the compensatory movement detection method proposed in this study is first compared with existing representative algorithms in terms of performance. Through multi-index quantitative evaluation, the advantages of the proposed method in the detection of compensatory movements after stroke are revealed. In addition, the control variable method is used to conduct ablation experiments on the model, analyze the contribution of each module to the performance, and verify the scientific nature of the architecture design.

### A. Evaluation Index

In this study, the experimental data of 15 stroke patients were divided into a training set and a test set at an 8:2 ratio. To evaluate the performance of the classification model, four key metrics were selected: Accuracy, Precision, Recall, and F1-score, for quantitative analysis. These metrics intuitively reveal the correspondence between the prediction results of the model classifier and the true labels, enabling a comprehensive and precise assessment of the model's performance. The specific calculation formulas are as follows:

$$Accuracy = \frac{TP+TN}{TP+TN+FP+FN}$$

$$Precision = \frac{TP}{TP+FP}$$

$$Recall = \frac{TP}{TP+FN}$$

$$F1 = \frac{2*Precision*Recall}{Precision+Recall}$$

Among them, TP, TN, FP, and FN represent True Positive, True Negative, False Positive, and False Negative of the classification labels respectively.

### B. Comparison Experiment

The performance of each model in detecting compensatory movements in the three rehabilitation movement tasks designed in this study is detailed in Table 2. The experimental data reveals that the detection accuracy of the methods based on traditional machine learning lies within the range of 0.6481 to 0.7901. Among them, the Support Vector Machine (SVM) attains the best performance by virtue of its superiority in nonlinear classification. However, the GCN-LSTM-ATT fusion model proposed in this research demonstrates remarkable performance advantages, with its detection accuracy reaching as high as 0.858. This outcome fully validates the effectiveness of the synergistic effect of the modeling capability of the Graph Convolutional Network (GCN) for the spatial topological structure of joints, the capturing ability of the Long Short-Term Memory network (LSTM) for temporal features, and the focusing ability of the Attention mechanism (ATT) for key time series features. It is capable of precisely capturing the dynamic topological relationships between joints in standard rehabilitation movements and their corresponding compensatory movements, significantly enhancing the recognition accuracy of the model for complex compensatory movement patterns.

Table 2 Performance of Each Model

| Model | Accuracy | Precision | Recall | F1-score |
|---|---|---|---|---|
| SVM | 0.7901 | 0.8267 | 0.7901 | 0.7887 |
| KNN | 0.6481 | 0.6562 | 0.6481 | 0.6459 |
| RF | 0.6852 | 0.7599 | 0.6852 | 0.6903 |
| GCN-LATM-ATT | 0.8580 | 0.8695 | 0.8580 | 0.8603 |

### C. Ablation Experiment

An ablation study is conducted on each module of the proposed GCN-LSTM-ATT using the real collected data to evaluate the impact of the three core components, namely GCN (serving as the baseline), LSTM, and ATT, on the model performance. Firstly, a spatiotemporal averaging operation is performed on the output features of GCN, and the result is input into the classifier, which is regarded as the baseline. The relevant experimental results are shown in Table 3.

As shown in Table 3, compared with the baseline GCN, GCN-LSTM-ATT has achieved a significant improvement in accuracy, verifying the great value of the proposed model in modeling the spatial topological structure of joints, capturing temporal features, and focusing on key temporal features in the task of detecting compensatory movements after stroke. In addition, compared with the baseline GCN, the GCN-LSTM model has also achieved a considerable improvement in accuracy, which validates the importance of combining spatial and temporal research for enhancing the accuracy of compensatory movement detection.

Table 3 Results of the Ablation Experiment

| Model | Accuracy | Precision | Recall | F1-score |
|---|---|---|---|---|
| GCN | 0.5679 | 0.6841 | 0.5679 | 0.5607 |
| GCN-LSTM | 0.8457 | 0.8592 | 0.8457 | 0.8462 |
| GCN-LATM-ATT | 0.8580 | 0.8695 | 0.8580 | 0.8603 |

## IV. CONCLUSION

In this study, a novel deep learning model is constructed based on skeleton data for the detection of compensatory movements after stroke. Firstly, three innovative rehabilitation movement tasks are designed, and the skeleton data of stroke patients during the execution of these tasks are collected, providing fundamental data support for subsequent research. Subsequently, a comparative experiment is conducted between the deep learning model proposed in this study and three existing machine learning algorithms. A compensatory movement detection model is established to achieve high-precision recognition of compensatory movements.

The experimental results show that the GCN-LSTM-ATT model significantly outperforms other comparative models in the task of detecting compensatory movements after stroke based on skeleton data. Although the scale of the current research dataset is relatively limited, this achievement provides important theoretical basis and technical reference for subsequent research. Meanwhile, it enhances the interpretability of the model in clinical applications.

Future research will focus on expanding the scale of the dataset, optimizing the model architecture, and further exploring the detection of various compensatory movement patterns in stroke patients. The aim is to develop more efficient and intelligent auxiliary diagnostic tools, effectively reducing the workload of clinicians and promoting the intelligent development of the field of stroke rehabilitation diagnosis and treatment.